\documentclass[conference]{IEEEtran}
\usepackage{cite}
\usepackage{graphicx}
\usepackage{amsmath}
\usepackage{amsmath}
\usepackage{amsfonts}
\usepackage{amssymb}
\usepackage{multirow}
\usepackage{xcolor}
\usepackage{float}
\usepackage{algorithm2e}
\usepackage{amsmath, amssymb, algpseudocode}
\usepackage{caption}
\usepackage{url}
\begin{document}
\bstctlcite{IEEEexample:BSTcontrol}
\RestyleAlgo{ruled} 
\title{Efficient Collaborations through Weight-Driven Coalition Dynamics in Federated Learning Systems}

\author{
\hspace{-0.75cm}
Mohammed El Hanjri$^1$, Hamza Reguieg$^2$, Adil Attiaoui$^1$,\\ Amine Abouaomar$^3$, Abdellatif Kobbane$^1$, Mohamed El Kamili$^2$\\
\\

$^1$ENSIAS, Mohammed V University in Rabat, Morocco\\
$^2$Higher School of Technology, Hassan II University in Casablanca, Morocco\\
$^3$School of Science and Engineering, Al Akhawayn University in Ifrane, Morocco.\\
\\
\vspace{0.1cm}
\textit {mohammed.elhanjri@um5r.ac.ma, hamza.reguieg-etu@etu.univh2c.ma, adil\_attiaoui@um5.ac.ma,}\\
\textit{a.abouaomar@aui.ma, abdellatif.kobbane@ensias.um5.ac.ma, m.elkamili@ieee.org}
}

\maketitle

\begin{abstract}
In the era of the Internet of Things (IoT), decentralized paradigms for machine learning are gaining prominence. In this paper, we introduce a federated learning model that capitalizes on the Euclidean distance between device model weights to assess their similarity and disparity. This is foundational for our system, directing the formation of coalitions among devices based on the closeness of their model weights. Furthermore, the concept of a barycenter, representing the average of model weights, helps in the aggregation of updates from multiple devices. We evaluate our approach using homogeneous and heterogeneous data distribution, comparing it against traditional federated learning averaging algorithm. Numerical results demonstrate its potential in offering structured, outperformed and communication-efficient model for IoT-based machine learning.\\

\end{abstract}
\begin{IEEEkeywords}
Federated Learning, IoT Devices, Euclidean Distance, Heterogeneity, Coalition Formation
\end{IEEEkeywords}

\IEEEpeerreviewmaketitle

\section{Introduction}
The Internet of Things (IoT) has ushered in a new era of connectivity, with billions of devices from wearables and smartphones to industrial sensors and smart home appliances interacting in complex networks. These devices continuously generate vast amounts of data, offering unprecedented opportunities for data-driven insights and machine learning applications \cite{cui2018survey, abouaomar2021deep}. However, the traditional paradigm of centralizing this data for processing and analysis is increasingly being challenged. Concerns related to bandwidth consumption, latency, privacy breaches, and the sheer volume of data make centralized approaches less feasible and efficient. This is where Federated Learning (FL) emerges as a promising alternative.

Federated Learning is a decentralized machine learning approach that allows devices, or "nodes", to train models on their local data without sharing the raw data itself. Instead of sending data to a central server, nodes process data locally and only transmit model updates or gradients to a central entity. This not only preserves user privacy by keeping sensitive data on the device but also reduces the bandwidth required for data transmission, making it particularly suitable for IoT scenarios where devices might be constrained by limited connectivity or power \cite{nguyen2021federated, khan2021federated, abouaomar2022federated}.

However, the adoption of FL in the IoT landscape is not without challenges. One of the most significant challenges is heterogeneity. In the vast and diverse world of IoT, devices differ widely in terms of computational capabilities, storage, power availability, and even the nature and distribution of the data they hold. For instance, data from a wearable health monitor will be vastly different from that of a smart thermostat. This non-IID (independent and identically distributed) nature of data across nodes can lead to disparities in local model updates, which, when aggregated at the central server, can adversely affect the global model's accuracy and convergence rate \cite{sattler2019robust, taik2024green}.

Moreover, the computational heterogeneity among devices means that while some devices can compute updates rapidly, others might lag, leading to asynchronous updates. This asynchronicity can introduce challenges in aggregating updates and ensuring that the global model remains current and relevant for all participating nodes. In essence, as the IoT ecosystem continues its exponential growth, the need for decentralized learning mechanisms like FL becomes increasingly evident. However, the inherent heterogeneity of this ecosystem poses unique challenges that need to be addressed to harness the full potential of FL \cite{abdellatif2022communication, elallid2023vehicles}. Through strategic approaches like coalition formation, there is potential to navigate these challenges and pave the way for more efficient, accurate, and privacy-preserving machine learning in the IoT domain.\\

One potential strategy to navigate the challenges posed by heterogeneity is the concept of coalition formation. By allowing devices to form strategic groups or "coalitions" based on similarities in data distribution, computational capabilities, or other criteria, it might be possible to create more homogeneous subgroups. These coalitions can then train local models that are more representative of their specific data distributions, leading to more accurate and robust global models when these local updates are aggregated.

The rest of this paper is organized as follows, with section II discussing related works. The system model is described in Section III, while Section IV presents the simulation and numerical findings. Section V serves as the paper's conclusion.\\

\section{Related Works}

The concept of Federated Learning has been extensively studied in recent years. A seminal work by researchers highlighted the challenges of communication in federated learning, emphasizing the need for structured and sketched updates to mitigate uplink communication costs and enhance efficiency. This work laid the foundation for subsequent studies that sought to integrate federated learning into various domains, such as internet of things. In this context, the decentralized training approach was underscored for its ability to maintain user data privacy, a critical concern in mobile applications. Furthermore, the applicability of federated learning was extended to the realm of the Internet of Things (IoT) \cite{10011632}. Comprehensive surveys in this domain emphasized the significance of decentralized learning in connected devices, addressing challenges and charting future directions for federated learning in IoT \cite{imteaj2021survey}.

The landscape of the Internet of Things (IoT) is rapidly evolving, with Machine Learning (ML) driving significant advancements in the realm of data communication for intelligent devices. Within this IoT framework, Federated Learning (FL) has emerged as a decentralized approach, with its application spanning across a myriad of IoT gadgets. While FL has gained considerable traction, many of its algorithms operate on the presumption that client nodes are trustworthy and collaborative during model training. However, real-world applications reveal that nodes can be deceptive or uncooperative. In this context, \cite{kabbaj2023distfl, el2023federated} introduces "DistFL", an innovative FL algorithm designed to thwart biased training by pinpointing malicious nodes during the training process, setting it apart from traditional FL methodologies. Concurrently, \cite{reguieg2023comparative} delves deeper into the FL paradigm, particularly contrasting the Federated Averaging (FedAvg) and Personalized Federated Averaging (Per-FedAvg) strategies. Their analysis accentuates the pivotal role data heterogeneity plays in the effectiveness of these techniques, especially when data is Non-Identically and Independently Distributed (Non-IID). Notably, in scenarios of pronounced data disparity, Per-FedAvg showcases enhanced resilience and performance. This highlights the need for tailored FL strategies in an IoT context, ensuring efficient and robust decentralized learning.

The fusion of Federated Learning and game theory has been a subject of interest in recent research. One of the notable works in this domain is the DIM-DS model \cite{chen2022dimds}, which introduces a dynamic incentive mechanism based on evolutionary game theory to model users' behavior in data sharing. The model also introduces "credibility coins" as a novel cryptocurrency for data-sharing transactions, emphasizing the importance of trust and incentives in federated systems.

Another significant contribution is the Towards Bilateral Client Selection in Federated Learning \cite{wehbi2022bilateral}, which proposes an intelligent client selection approach for federated learning on IoT devices using matching game theory. This approach surpasses traditional federated learning client selection methods by optimizing both the revenues of client devices and the accuracy of the global federated learning model.
The FedGame framework \cite{houda2022fedgame} is another noteworthy mention, designed to secure Industrial Internet of Things (IIoT) applications leveraging FL. This framework evaluates its accuracy against centralized ML/DL schemes, emphasizing not only the privacy of industrial systems but also its efficiency in providing required resources to counter IIoT attacks.

Federated Learning facilitates the joint training of machine learning models by enabling collaboration between an aggregation server and various clients, all while preserving data privacy. However, this privacy-centric approach renders the FL structure susceptible to potentially harmful model updates, primarily since the server perceives clients as opaque entities. Such a setup, compounded by vast heterogeneous data, magnifies security vulnerabilities and may degrade model performance. Despite the significance, the ramifications of client selection and data heterogeneity on FL's resilience are often understated. In \cite{10239107}, the authors introduced the Incentive Design for Heterogeneous Client Selection (IHCS) to bolster performance and mitigate security concerns. IHCS innovatively leverages cooperative game theory and dynamic client clustering based on data heterogeneity. By assigning a unique Shapley Value-derived recognition to each client, it refines client participation probabilities. Furthermore, our incorporation of a heterogeneity-focused clustering technique, known as HIC, counters the adverse effects of data diversity, enhancing the overall contributions from client.

The proliferation of IoT devices has unveiled a vast reservoir of IoT data. ML models leveraging this data are instrumental in monitoring vital environmental parameters such as air quality and climate change. Traditional centralized ML models, however, consolidate data from all clients at one central location, compromising user privacy. Federated Learning (FL), an alternative, prioritizes user privacy by training models on dispersed data sets. Yet, FL grapples with a heterogeneity challenge, often favoring some data sources at the expense of others. To address this, authors in \cite{9134408} introduce a central server, fortified by reinforcement learning (RL), that discerns data disparities. This server, within the FL framework, gauges collaborative benefits by interpreting patterns in diverse client feedback. By iteratively adjusting client weights, it strives to achieve a coalition of clients that delivers near-optimal performance.

\section{System model}

The proliferation of the Internet of Things has ushered in a new era of interconnected devices, from smart thermostats in homes to intricate sensors in industrial settings. These devices, while constantly collecting and processing data, are often constrained in terms of computational power and memory. Our model introduces a decentralized paradigm, leveraging federated learning, tailored specifically for the IoT ecosystem.

Central to our approach is the use of the Euclidean distance between the weights of local device models. In the vast expanse of the IoT, where devices are continuously learning and adapting, each device's model weights can be conceptualized as vectors in a high-dimensional space. The Euclidean distance between these vectors serves as a metric to measure the similarity or disparity between two models. A smaller distance implies that the models are close, while a larger distance denotes significant differences. This metric is foundational for our system, guiding the formation of coalitions among devices based on the proximity of their model weights.

To further refine the aggregation of updates from multiple devices, we employ the concept of a barycenter. In this context, the barycenter symbolizes the average of model weights across a subset of devices. It acts as a central point in the weight space, encapsulating the collective learning of a group of devices. This is especially pivotal when performing federated averaging across multiple devices in the IoT network.

Our model ensures that devices, based on the similarity of their local models, are grouped into coalitions. These coalitions then collaboratively contribute to a global model, minimizing communication overheads, making it a promising approach for the future of IoT-based machine learning.

\subsection{Euclidean distance between weights}

In the context of federated learning, the weights of a model are typically represented as vectors (or matrices, but they can be reshaped into vectors for the purpose of distance computation). The Euclidean distance between the weights of two users' models can be defined as the distance between these two vectors in the Euclidean space.\\

Given two users with model weights \( \boldsymbol{\omega}_1 \) and \( \boldsymbol{\omega}_2 \), where each \( \boldsymbol{\omega} \) is a vector in \( \mathbb{R}^n \), the Euclidean distance \( d \) between these weights is defined as:

\[
d(\boldsymbol{\omega}_1, \boldsymbol{\omega}_2) = \sqrt{\sum_{i=1}^{n} (\omega_{1_i} - \omega_{2_i})^2}
\]

Where:
\begin{itemize}
    \item \( n \) is the number of elements (or dimensions) in the weight vectors.
    \item \( \omega_{1_i} \) and \( \omega_{2_i} \) are the i-th elements of the weight vectors \( \boldsymbol{\omega}_1 \) and \( \boldsymbol{\omega}_2 \) respectively.\\
\end{itemize}

In simpler terms, the Euclidean distance measures the "straight-line" distance between two points in Euclidean space. In the context of federated learning, this distance can give an indication of how different two users' local models are from each other. If the distance is small, it suggests that the two users' models are similar, while a large distance indicates that the models are quite different.\\

\subsection{Barycenter of a subset of users}

In the context of federated learning, the barycenter of the weights of a subset of users can be thought of as the average of their model weights. This concept is often used when aggregating updates from multiple devices in federated averaging.\\

Given a subset \( S \) of \( m \) users, where each user \( u_i \) has model weights \( \boldsymbol{\omega}_{i} \) (represented as vectors in \( \mathbb{R}^n \)), the barycenter $\textbf{b} $ of their weights is defined as:

\[
\boldsymbol{b} = \frac{1}{m} \sum_{i=1}^{m} \boldsymbol{\omega}_{i}
\]

This means that for each dimension (or element) of the weight vector, you compute the average value across all users in the subset.\\

As with the coalition, this definition assumes that each user in the subset has equal importance. If users have different importance or weights, the barycenter would be a weighted average of their model weights.\\

\subsection{Model Description}

\begin{figure}[h]
    \center
    \includegraphics[width=0.5\textwidth]{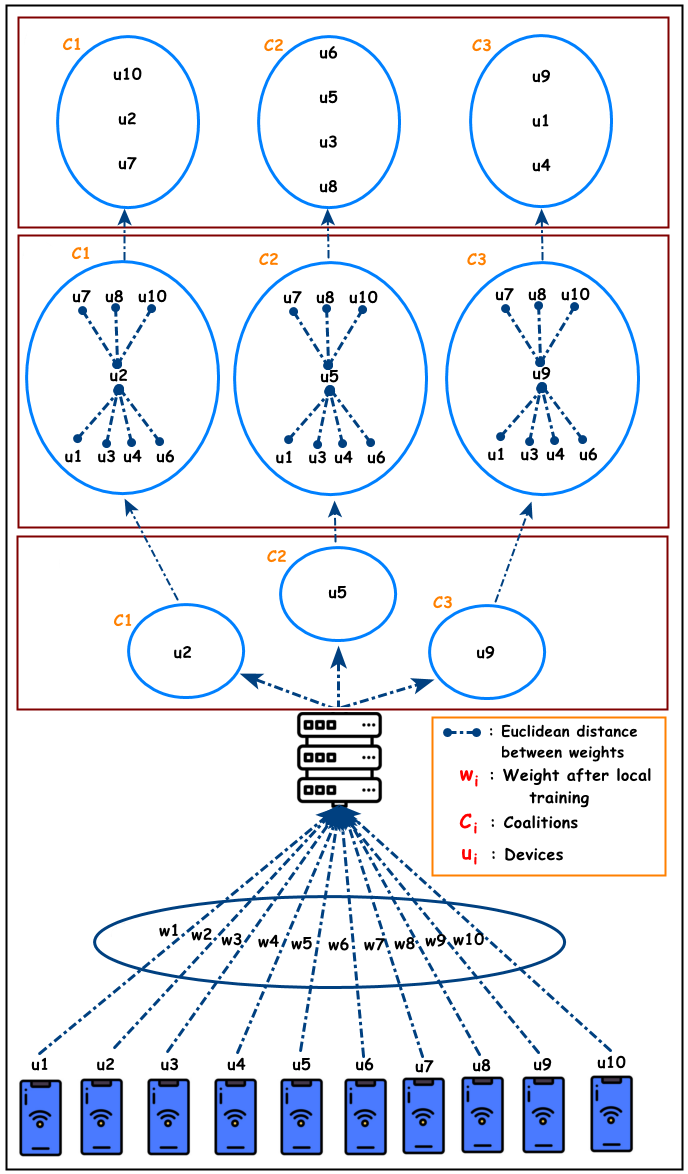}
    \caption{An illustrative overview of the different stages of coalition formation in one global training round.}
    \label{archi}
\end{figure}

\subsection*{\textbf{Step I: Coalition initialization}}
We start by selecting a subset of 10 devices, $U = \{ u_i \}_{i\in \{1,2, ..., 10\}} $, for our initial study. Each device from $U$ associated with $\omega_i^r$ as the weight after local training in the round $r$, We then initialize three empty coalitions $C_j, j \in \{1, 2, 3\}$.\\

We randomly select three devices from $U$ as the initial centers of these coalitions in round $r= 0$ noted $v_j^0, j \in \{1, 2, 3\}$. The criterion for selection is that the distance between the model weights of any two chosen devices should be non-zero, ensuring diversity, i.e.
$$d(\omega_{v_j^0}, \omega_{v_{j'}^0})>0 \quad \forall j,j' \in \{1,2,3\} \quad and \quad j \neq j'$$

\subsection*{\textbf{Step II: Device assignment to the corresponding coalition}} 

Each device in our subset is then assigned to the coalition whose center is closest to it, based on the Euclidean distance between their model weights. This step ensures that devices with similar data patterns or environmental conditions are grouped together, potentially leading to more accurate local models.

\begin{itemize}
 
\item For each user $u_i \in U$, We calculate the distance between the user $u_i$ and each coalition center $v_j^r,$ i.e. $d(\omega_i^r, \omega_{v_j^r}), \quad \forall u_i \in U \setminus \{v_j^r\}_{j\in {\{1,2,3\}}} $

\item Assign $u_i$ to the corresponding coalition at the nearest center; i.e. $u_i$ joins the coalition verifying:
$$ min \quad d(\omega_i^r, \omega_{v_j^r}) , \quad j\in \{1,2,3\}$$
\end{itemize}

\subsection*{\textbf{Step III: Calculation of each coalition's barycenter and updating of coalition centers}}

\subsection*{\textbf{Calculation of each coalition's barycenters}}
\begin{itemize}
 
\item After assigning each user to a coalition, we recalculate the center (barycenter) of each coalition.
\item For a coalition \( C_j \), its barycenter is defined as: 
\[ b_j^r = \frac{1}{|C_j|} \sum_{u_i \in C_j} \omega_i^r \]
where \( |C_j| \) is the number of users in the coalition \( C_j \).\\

\end{itemize}

\subsubsection*{\textbf{New coalition centers update}}
The new center $v_j^{r+1}$ of each coalition $C_j$ for the next iteration of global model is the user verifying 
$$  min \quad d(\omega_i^r, b_j^r)$$

\subsubsection*{\textbf{Step IV: Global model aggregation}}

The central server, which could be a cloud-based platform or a local gateway, aggregates the barycenters of all coalitions to form a global model.

The global server model is the average of the barycenters of all the coalitions $C_j$, 

$$ \theta^{(r)} = \frac{1}{3}\sum_{j=1}^3  b_j^{r-1}$$

\begin{algorithm}
Initialize global model parameters \( \theta^{(0)} \) ;\\
 Initialize three empty coalitions: \( C_j, j\in \{1, 2, 3\} \) ;\\
 \( \omega_{i}^0 \leftarrow \text { ClientUpdate }\left(u_i, \theta^{(0)} \right) \);\\
\While{\( d(\omega_i^0,\omega_k^0)_{i \neq k} > 0 \) }{
Choose three users randomly noted \( v_j^0 \) as initial centers of  \( C_j \) respectively;\\
break\;
}
\For{each round \( r=1,2, \ldots \) }{
    \ForEach{user \( u_i  \in \) \( U\setminus\{  v_j^{r-1} \}_{j\in \{1, 2, 3\}} \)}{
        Calculate \( d(\omega_i^{r-1},\omega_{v_j}^{r-1})  \) \( \forall j\in \{1, 2, 3\} \);\\
        Assign \( u_i \) to the coalition \(C_j\) verifying \\ \( min \quad d(\omega_i^{r-1},\omega_{v_j}^{r-1})  \) \( \forall  j\in \{1, 2, 3\} \) ;
    }
    \Return \( C_j \) ;\\
    \ForEach{coalition \( C_j \) where \( j \in \{1, 2, 3\} \)}{
        Calculate the barycenter of the weights \\ \( b_j^{r-1} = \frac{1}{|C_j|} \sum_{u_i \in C_j} \omega_i^{r-1} \) ;\\
          Assign \( u_i \) as the new center \(v_j^r\) verifying \\ \( min \quad d(\omega_i^{r-1},b_j^{r-1})  \) \(\forall  j \in \{1,2,3 \} \) ;\\
    }
      \(  \theta^{(r)} = \frac{1}{3} \sum_{j=1}^3 b_j^{r-1} \);\\
      \( \omega_{i}^r \leftarrow \text { ClientUpdate } \left(u_i, \theta^{(r)} \right) \) ;\\
}

\caption{Federated Learning with Coalition Formation based on Euclidean Distance between Weights}
\end{algorithm}


\section{Simulation and Numerical Results}

\subsection{Simulation Setup}

All experiments were conducted on a laptop powered by an Intel(R) Core(TM) i5-1035G1 CPU @ 1.00GHz. The hardware specifications included 4 cores and 8 logical processors, supported by 12GB RAM. The base frequency of the processor was 1190 MHz.\\
Across all experiments, 10 users (or clients) participated, each receiving a distinct portion of the dataset. The primary objective was to assess the disparity in performance between the traditional FedAvg algorithm and our proposed algorithm under different data distribution scenarios: IID, heterogeneous, and highly heterogeneous manners.\\

\subsection{Dataset}

The MNIST dataset was the primary focus of our study, renowned for its role in handwritten digit classification tasks. This dataset comprises 70,000 grayscale images of handwritten digits (0 to 9) with dimensions of 28 × 28 pixels. For the purpose of training and evaluation, the dataset was divided into 60,000 images for training and 10,000 for testing. The data was further segmented into 10 parts, each containing 6,000 samples. These samples were distributed among clients in an IID manner as well as heterogenous and highly heterogenous manner, ensuring each client had 600 samples for every class.\\

\subsection{Algorithm Implementation}

The experiment was implemented using PyTorch, a popular deep learning framework. The primary objective was to compare the performance of the FedAvg algorithm with the newly proposed algorithm. A Convolutional Neural Network (CNN) model was employed for the image classification task.\\

\subsection{Model Architecture}

The CNN model used was designed specifically for the MNIST dataset. It consisted of:

Two convolutional layers (conv1 and conv2) with respective output channels of 32 and 64, both using a 5x5 kernel size and ReLU activation.
Two max-pooling layers (pool1 and pool2) with a 2x2 kernel and a stride of 2.
Two fully connected layers (fc1 and fc2) with 512 neurons and 10 neurons respectively. The latter corresponds to the 10 classes of the MNIST dataset.
The forward pass involved reshaping the input tensor, passing it through convolutional layers, max-pooling, and then the fully connected layers to produce class logits.
This model is a typical representation of CNN architectures used for image classification tasks, leveraging convolutional layers for feature extraction, max-pooling for spatial reduction, and fully connected layers for classification.\\

\subsection{Parameters and Configuration}

Throughout the experiment, each communication round saw devices executing 5 local training epochs with a batch size of 10. This was done to assess the effectiveness of both the FedAvg and our proposed algorithm in the learning process. The model was trained using the SGD optimizer.\\

\subsection{Discussion}
In the exploration of the performance metrics across three distinct data scenarios: homogeneous, moderately heterogeneous, and highly heterogeneous data. The disparities between Federated Learning methodologies employing standard FedAvg and the proposed FL with Coalitions become evident.\\

\begin{figure}[h]
    \center 
    \includegraphics[width=0.5\textwidth]{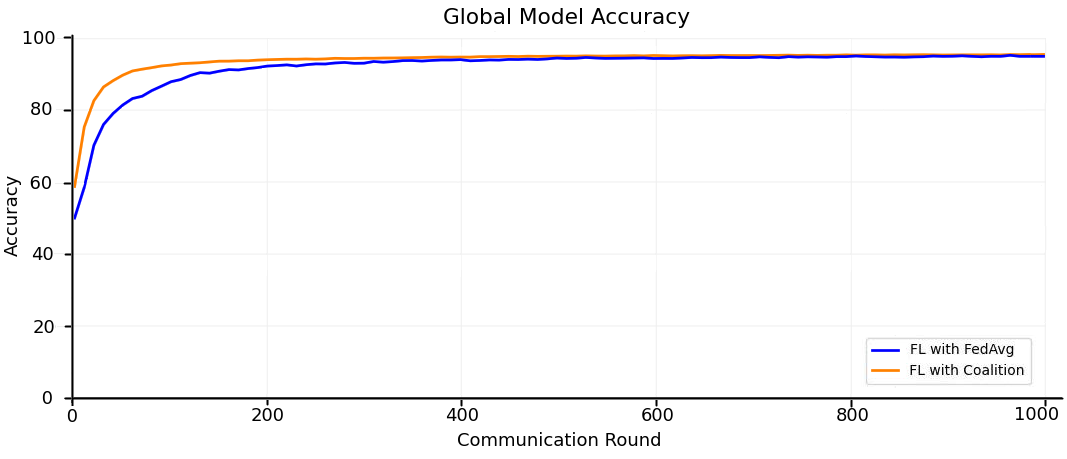}
    \caption{Accuracy of FedAvg and the proposed approach in case of homogeneous data distribution}
    \label{iid}
\end{figure}

With the homogeneous dataset shown in Fig. \ref{iid}, both methodologies exhibited impressive performance metrics, particularly in terms of model accuracy. Convergence was achieved swiftly, revealing stability in accuracy as the communication rounds advanced. The uniform nature of the dataset clearly favors decentralized learning strategies due to consistent data distribution across nodes. Yet, even in these conditions, the slight edge FL with Coalitions held demonstrated its potential robustness in the aggregation process, possibly alluding to the benefits of coalition-driven dynamics.\\

\begin{figure}[h]
    \center
    \includegraphics[width=0.5\textwidth]{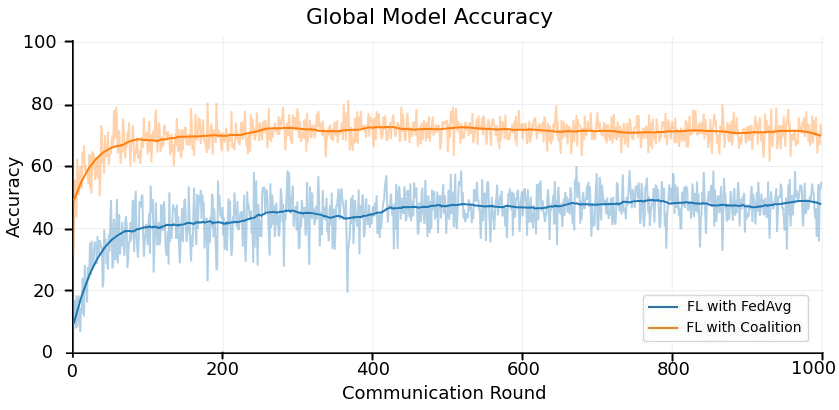}
    \caption{Accuracy of FedAvg and the proposed approach in case of heterogeneous data distribution}
    \label{0.5}
\end{figure}

Transitioning to the moderately heterogeneous data scenario in Fig. \ref{0.5} presented more pronounced variances between the two methodologies. The standard FedAvg displayed higher fluctuations in accuracy during initial communication rounds, gradually achieving stability. In contrast, FL with Coalitions showcased a consistent upward trajectory in accuracy, suggesting its adeptness in managing data variance. The relative stability of this coalition-based approach highlights its proficiency in addressing the challenges of data disparity, potentially by capitalizing on the collective strength of nodes with similar data.\\

\begin{figure}[h]
    \center
    \includegraphics[width=\linewidth]{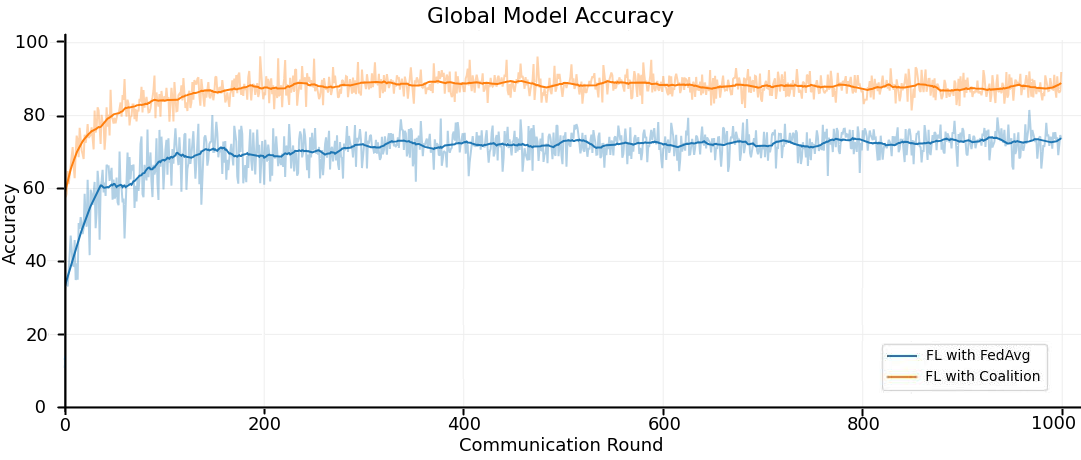}
    \caption{Accuracy of FedAvg and the proposed approach in case of highly heterogeneous data distribution}
    \label{0.1}
\end{figure}

The most revealing contrast came with the evaluation of highly heterogeneous data Fig.\ref{0.1}. Both methodologies grappled with the data's disparate nature, but it was the FL with Coalitions that maintained controlled accuracy ascents, demonstrating its resilience. Meanwhile, the standard FedAvg showcased pronounced volatility, reflecting its limitations in such environments. This superior performance of FL with Coalitions indicates its adaptability and suggests that its mechanism, possibly its nuanced approach to clustering similar nodes, mitigates the impact of pronounced data disparities.\\

\section{Conclusion}
Our exploration into the field of federated learning for the IoT domain led to the design of a decentralized learning model leveraging the Euclidean distance between model weights of individual devices. The results highlight the efficiency of our proposed model in forming coalitions among devices based on the similarity of their model weights. The use of a barycenter to represent the average model weights has also shown to be effective for aggregating updates from several devices. By conducting our experiments, we provided a comprehensive comparison between the conventional FedAvg algorithm and our federated learning approach using coalition formation . Our findings underscore the significance of utilizing such techniques in the IoT context, minimized communication overheads, and efficient learning.\\

\section*{Acknowledgement}
This research was supported by the Alkhawarizmi AI Project (grant number: Alkhawarizmi/2020/34).

\bibliographystyle{IEEEtran}
\bibliography{refs}

\end{document}